\documentclass[12pt]{article}
\usepackage{amssymb}
\usepackage{graphicx}
\usepackage{subfigure}
\usepackage{indentfirst}
\usepackage[square, super, comma, sort&compress]{natbib} %style of ref
\usepackage[a4paper,left=3.2cm, right=3.2cm, top=2cm, bottom=3cm]{geometry}
\usepackage{titlesec}
\titleformat*{\section}{\large\bfseries}

\usepackage{hyperref}
\usepackage{amsmath}
\usepackage{amsfonts}
\usepackage{float}
\usepackage{caption}

\usepackage{setspace} % row spacing 1.5
\begin{document}

%--------------------------------------------------

\begin{center}

	{\large \textbf{Advanced Computing and Related Applications Leveraging Brain-inspired Spiking Neural Networks}}\\ \vspace{3em}

	{\bf Corresponding Author:} Nicole L. Yien\\ \vspace{1em}

    {\bf Secondary Authors:} Lyuyang Sima, Joseph Bucukovski, Erwan Carlson\\ \vspace{1em}
 
	September 2023 \vspace{2em}
 
\end{center}

%--------------------------------------------------

\begin{abstract}

In the rapid evolution of next-generation brain-inspired artificial intelligence and increasingly sophisticated electromagnetic environment, the most bionic characteristics and anti-interference performance of spiking neural networks show great potential in terms of computational speed, real-time information processing, and spatio-temporal information processing. Data processing. Spiking neural network is one of the cores of brain-like artificial intelligence, which realizes brain-like computing by simulating the structure and information transfer mode of biological neural networks. This paper summarizes the strengths, weaknesses and applicability of five neuronal models and analyzes the characteristics of five network topologies; then reviews the spiking neural network algorithms and summarizes the unsupervised learning algorithms based on synaptic plasticity rules and four types of supervised learning algorithms from the perspectives of unsupervised learning and supervised learning; finally focuses on the review of brain-like neuromorphic chips under research at home and abroad. This paper is intended to provide learning concepts and research orientations for the peers who are new to the research field of spiking neural networks through systematic summaries.

\end{abstract}	

\vspace{5mm}

%--------------------------------------------------

\section{Introduction}

Humans being, widely considered to be the most intelligent creatures on this blue planet, possess complicated biological neural networks within their brains that exhibit remarkable efficiency and robustness in handling various tasks, from simple reflex actions to advanced problem-solving and decision-making. It is generally believed that the energy consumed by the human brain every day, expressed in the form of electrical energy, is a mere 25 watts. This has inspired many scholars to devote themselves to brain science research, delving into the workings of biological neural networks in the human brain and simulating the way the brain processes and remembers information for pattern recognition and intelligent control \cite{widrow1994neural}. In last century, the ``Human Brain Project'' was officially launched in the United States of America. On April 2, 2013, USA President Obama announced the launch of the Brain Research through Advancing Innovative Neurotechnologies (BRAIN) Initiative. Subsequently, the European Union’s Brain Science Program (Human Brain Project HBP) and Japan’s Brain Science Program (Brain Mapping by Integrated Neurotechnologies for Disease Studies MINDS) were also launched. In 2018, the People's Republic of China officially released its Brain Project. China’s Brain Project aims to explore the cognitive principles of the brain and gradually apply brain science principles to medical diagnosis and brain-like intelligence. On one hand, it explores treatment plans for brain diseases and develops medical equipment; on the other hand, it develops new technologies related to brain-like artificial intelligence \cite{lindsay2021convolutional}

The human brain is an incredibly complex organ that is often compared to a vast network formed by nearly 100 billion neurons and their synaptic connections which communicate with each other through electrical and chemical signals, allowing us to think, feel, and then interact with the world around us. Like a tree with branches reaching out to receive information from other neurons, neurons in the network therefore receive information through dendrites and then pass it on to the next neuron through structures such as axons and synapses. Artificial neural networks (ANNs) mimic the process of information transmission in the human brain to perform calculations and are applied in many fields of technology and life, including speech processing, computer vision, and natural language processing. However, traditional ANNs still differ significantly from real biological neural networks. The inputs and outputs of traditional ANNs are real numbers, while the form of information transmission in the human brain is discrete action potentials or pulses. The spiking neural network (SNN) proposed by Maass in 1997 fully mimics biological neural networks by transmitting information in the form of pulses \cite{maass1997networks}. Essentially binary events, pulses play an important role in improving processing efficiency and reducing energy consumption. However, the discreteness of information transmission makes the implementation of SNN algorithms relatively difficult. Therefore, some scholars have designed learning algorithms based on the principles and characteristics of SNNs; while others have devoted themselves to applying existing ANN algorithm concepts more effectively to SNNs. Traditional ANN algorithms have hardware acceleration platforms such as central processing units (CPUs) and graphics processing units (GPUs). Efficient application of SNN algorithms also requires hardware support. In recent years, many research institutions and companies have achieved hardware results for SNNs. Brain-like neuromorphic chips have emerged and have great development prospects, with the potential to truly achieve brain-like artificial intelligence. In the near decade, artificial neural networks have also made great progress in human production life. People have applied these neural networks to the prediction of stock market, weather, electricity consumption and other application scenarios, and also applied neural algorithms to optical structure optimization \cite{peurifoy2018nanophotonic, liu2018training}, sensor demodulation and calibration \cite{saracoglu2008artificial, liehr2019real, kowarik2020fiber}, communication \cite{eriksson2017applying, ibnkahla2000applications, chen2019artificial}, etc., which greatly enriched the use of neural networks. Meanwhile, there are also researchers working on accelerating neural network algorithms at the hardware level and developing new specialized chips. The demand for high performance, high energy efficiency, and greater bandwidth in neuromorphic computing is endless. As the exponential growth of electronic transistors marked by Moore’s Law gradually approaches its physical limit, traditional silicon-based electronic components have reached a bottleneck. More new components have been proposed to try to meet the needs of neuromorphic computing, such as photonic computing chips \cite{cheng2017chip, van2017advances, photonics9100698, shastri2021photonics, antonik2019human, mao2019photonic, katumba2018low}, memristors\cite{huh2020memristors, prezioso2015training, jeong2016memristors}, phase change memory (PCM)\cite{lee2009architecting, salinga2018monatomic, raoux2014phase, fong2017phase}, nanoelectronics-spintronics-assisted computing devices \cite{li2015spintronic, li2018spin, torrejon2017neuromorphic, parkin2008magnetic, luo2018reconfigurable, grollier2020neuromorphic, hong2019demonstration, markovic2020physics}. These are all innovative memories with high processing speed, huge storage capacity, and good long-term stability that can better perform efficient neuromorphic computing.

This article reviews SNNs from the perspective of their good biomimicry, high efficiency, and low energy consumption, from neuron models and network topologies to learning algorithms and SNN brain-like neural chips. Like a map guiding the search for models and networks with more biological characteristics, it first summarizes the advantages and disadvantages of five spiking neuron models and the characteristics of five SNN network topologies. Then, from the perspective of learning methods, it reviews unsupervised learning algorithms as well as four types of supervised learning algorithms based on synaptic plasticity, backpropagation, convolution, and ANN weight conversion to SNN. Finally, it reviews SNN brain-like neuromorphic chips from the two circuit structures of analog-digital hybrid circuits and digital circuits.

%--------------------------------------------------

\section{Spiking Neural Network}
\subsection{Spiking Neuron Model}
The most basic element in an SNN is named the spiking neuron (SN), the two decisive variables of which are the membrane potential and the activation threshold. Whether a neuron fires or not is closely related to these two variables. If the membrane potential of a neuron in the network reaches the activation threshold, it will emit a spike signal that is transmitted to the next neuron through synapses. A large number of neurons work together to form a network for systematic learning. The most commonly applied models in SNN network construction are the Hodgkin-Huxley model (HH), the integrate-and-fire model (IF), the leaky integrate-and-fire model (LIF), the spike response model (SRM), and the Izhikevich model. The HH model can accurately represent the dynamic changes of ion channels, but it requires separate modeling of sodium, potassium, and leakage channels, and its expression is complex and its calculation process is cumbersome. The LIF model is currently the most commonly used model in SNN networks. Although it ignores the dynamic changes of ion channels and only reflects the macroscopic changes of membrane potential, its calculation is simple.

In addition to these five commonly used spiking neuron models, in recent years some scholars have innovatively combined biological mechanisms and mathematical properties to explore spiking neuron models that have both biological and computational characteristics. For example, Zuo et al. proposed a probabilistic spike response model from a probabilistic perspective. The firing mode of this model does not depend on the difference between the threshold and membrane voltage, nor does it depend on the shape of the spike. Instead, it reconstructs the relationship between membrane voltage and neuronal firing probability and transmits information in a probabilistic manner. On the other hand, spintronics is a promising platform for solid-state device technologies, especially for neural computing applications \cite{wolf2001sp, kou2014scale, hong2020dual, zhao2015sp, clemente2012magnetic}. It offers fast and virtually infinite information writing operation with standard CMOS-compatible voltages and can store its state without power. MRAMs utilizing spin-transfer torque-induced magnetization switching are about to hit the mass market \cite{prenat2015ultra}. These features hold promise for high-performance and low-power artificial neural networks that are adaptive and robust. Spintronics devices can represent digital information as their magnetization direction, but can also deal with analog information through their magnetic domain structures, providing opportunities for spintronic synapses in artificial neural networks. In summary, spintronics offers an attractive platform for high-performance, low-power, and adaptive artificial neural networks as well as logic computation \cite{li2018novel}. For spintronics devices to be feasible, an efficient scheme for electrically controlling magnetization is necessary. One leading example is STT-induced magnetization switching in magnetic tunnel junctions, which has enabled the successful development of STT-MRAMs. Recent studies have also shown that spin-orbit torque (SOT) provides a promising scheme for inducing magnetization switching and random number creating \cite{chen2018binary}. SOT arises when an in-plane current is introduced to magnetic crystals with noncentrosymmetry or magnetic heterostructures with broken space inversion symmetry and sizable spin-orbit interaction \cite{zhang2019Aspin, zhang2019spin, song2019spin}. The origin of SOT in heterostructures is still a subject of debate and may vary between systems.

\subsection{Topological Structure of Spiking Neural Network}
Multiple spiking neurons are connected through synapses to form large-scale SNN networks, and different connection methods determine the type of SNN network topological structure. Similar to traditional ANNs, SNN network topological structures can be divided into feedforward SNNs, recursive SNNs, recurrent SNNs, evolutionary SNNs, and hybrid SNNs. Feedforward SNNs consist of an input layer, a hidden layer, and an output layer. Each layer is composed of one or more neurons arranged in a row. Neurons are connected through multiple synapses with dynamically adjustable weights to transmit signals from the input layer to the next layer. Recursive SNNs and recurrent SNs both contain feedback loops and better reflect the connections between real neurons, but they increase the difficulty in algorithm design. Evolutionary SNNs have the characteristics of adaptability and self-organization and can dynamically adjust the number of neurons according to the characteristics of the samples. Hybrid SNNs have diverse local structural types, including feedforward and recursive types. At present, considering the complexity of topological structure implementation and the update iteration speed of network parameters, feedforward SNNs are most commonly used. Moreover, the model construction of SNN network topological structure can affect the choice of SNN learning algorithms.

\section{Learning algorithm of Spiking Neural Network}
After the network is constructed, SNN learning algorithms are required to train the data. According to the implementation form of SMN learning algorithms, they are able to be divided into two categories: unsupervised learning algorithms and supervised learning algorithms. The core of unsupervised learning algorithms is the spike-timing-dependent plasticity rule (STDP); supervised learning algorithms can be summarized into two categories according to their design ideas. One category revolves around the STDP rule, and the other uses ideas from ANs such as backpropagation (BP) and convolution for direct or indirect training. Unsupervised learning algorithms using the STDP rule can effectively reflect biological characteristics, but STDP has local characteristics. Each layer focuses on adapting to the output of the previous layer and cannot coordinate the entire network. It is not suitable for multi-layer structures and has a low classification accuracy. Many scholars have solved classification problems by modifying STDP learning rules, combining convolutional networks to extract deep features, and adding supervised learning tasks during training, thus producing a category of supervised learning algorithms based on STDP learning rules. Another category of supervised learning algorithms is based on AN ideas. The difficulty of direct training algorithms based on BP ideas lies in solving the non-differentiability of neuron equations, gradient explosion, and overfitting problems. Direct training algorithms based on convolutional ideas mainly focus on the selection and optimization of convolutional kernels. Indirect training algorithms based on ANNs first train ANs and then convert them to SNNs through normalization and other methods. In this section, we will provide a detailed overview of unsupervised learning algorithms and supervised learning algorithms

\subsection{unsupervised learning algorithms}
Researchers such as Song et al. proposed the STDP learning rule based on the learning rule proposed by Heobian. The STDP learning rule adjusts the strength of connections between neurons according to the order in which neurons learn. For any two neurons, if the presynaptic neuron fires earlier than the postsynaptic neuron, the connection strength between the neurons increases; otherwise, the connection strength between the neurons decreases. Studies have shown that shallow SNNs with unsupervised STDP algorithms as their core perform far worse in classification tasks than traditional ANs such as convolutional neural networks (CNNs). CNNs are widely used in various fields such as pattern recognition and image classification, especially deep CNNs perform well in extracting key features. However, the weight connection method in CNNs has no biological basis. Therefore, scholars tend to combine STDP rules and CNNs to more comprehensively leverage the advantages of CNNs in computational accuracy and SNNs in computational efficiency. For example, Lee et al. proposed a deep spiking convolutional neural network (SpICNN) that uses an unsupervised method based on STDP rules to train two convolutional layers. Sinivasan et al. proposed a probabilistic learning algorithm based on STDP rules called hybrid-STDP (HB-STDP), which combines STDP and anti-STDP learning mechanisms to train a residual stochastic multilayer convolutional spiking neural network (ReStoCNet) composed of binary kernels in a hierarchical unsupervised learning manner. Overall, SNN unsupervised learning algorithms that combine convolution and STDP learning rules can leverage the characteristics of both CNNs and SNNs and have both computational efficiency and biomimicry.

\subsection{supervised learning algorithms}
\subsubsection{supervised learning algorithms based on STDP}
The STDP learning rule was first natural one applied to unsupervised learning algorithms \cite{kheradpisheh2018stdp, serrano2013stdp}, but unsupervised learning algorithms are only suitable for clustering problems and have limited adaptability. Therefore, researchers such as Ponulak modified the STDP learning rule and proposed the remote supervised method (ReSuMe), which combines the STDP rule with remote supervision to minimize the difference between output and target spikes without calculating gradients. Taherktan and others used STDP, anti-STDP, and delay learning rules to learn the parameters of the hidden and output layers in parallel, allowing weight and delay learning to interact and greatly improving the accuracy of the algorithm while also being more biologically plausible. In addition, some researchers have proposed reward-modulated STDP (RR-STDP) rules inspired by the role of neuromodulators such as dopamine and acetylcholine in STDP regulation. Mozalar and others applied both STDP and P-STDP rules to deep convolutional SNNs, using STDP in the first layer and R-STDP in the latter two layers, achieving a recognition rate of 97.2\% on the MNIST dataset. Supervised learning algorithms based on the STDP rule have improved the accuracy of classification tasks while maintaining biological plausibility. 

\subsubsection{Direct and Indirect Learning Algorithm Based on ANN}
Direct and indirect learning algorithm based on ANN idea

(1) Direct training algorithm based on backpropagation: Backpropagation and gradient descent are important means to achieve optimization in neural networks. Using appropriate backpropagation and gradient descent ideas in weight update can solve the non-differentiability problem of SNN. Bohte et al first applied the backpropagation and gradient descent ideas to SNN, and introduced the supervised learning algorithm SplkeProp into the backpropagation process, calculated the gradient descent according to the principle of minimum error, and updated the synaptic weights to obtain the optimal solution. The direction propagation algorithm has the background of traditional learning algorithms, so there are problems in many aspects, such as: the gradient problem will make the learning process inefficient;; the global error information incorporated in the learning process lacks biological support; in order to improve the accuracy of the algorithm, it needs Increase the number of hidden layers, but too many hidden layers will cause overfitting problems, making the learning process not robust to interference. Hong et al. proposed an improved SpikePcp learning algorithm and designed a pulse gradient The threshold rule is used to solve the gradient explosion problem in SNN training. In order to control the network activity during the training process, the adjustment rules of pulse emission rate and connection weight are also added. In addition, scholars consider various influencing factors in backpropagation , such as axonal delay, local propagation form 81, space-time domain synergy R31, macro-micro diversity 2l, approximate activation function 8~, and agent gradient M\%, etc., gradually improve the classification accuracy of the algorithm. The most known Under optimal conditions, the classification accuracy rate on the MNIST data set can reach 99.49\%. 

(2) Direct training algorithm based on convolution: WicowrHif is now is one of the commonly used weight adjustment algorithms in linear neural networks, which is suitable for analog signals, but the pulse sequence is a discrete signal, and the Widrow-Hof rule cannot be directly applied. Therefore, scholars use the idea of convolution to process discrete data to make it have continuous features. Usually, by adding a convolution kernel, the elements of the impulse vector are correspondingly transformed into continuous functions: The spike pattern joint neuron algorithm proposed by Mohemmed et al.  converts the spike sequence (input spike sequence, neuron target and actual output spike sequence) into a continuous function signal through a kernel function, and then The Wcrow-Hof rule is used to adjust the synaptic weight. The precise spike-driven plasticity algorithm proposed by Y et al. (precise-spike-drven, PSD) only uses the kernel function to convert the input spike sequence into a convolution signal, and the target output pulse and The error between the actual output pulses drives the synapse to achieve self-adaptation. The core of the algorithm based on the convolution idea focuses on the selection of the convolution kernel. Lin Xianghong et al. The spiking training convolution kernel learning rule (spike ran kernel earning ruie.STKLR), tested various kernel functions in the STKLR algorithm. 

(3) ANN-based indirect training algorithm: The development of AN has entered a mature stage and is widely used in image recognition, target recognition, unmanned driving, bioinformatics and other fields. If the mature algorithm of traditional ANN is indirectly applied to SNV, good results may be obtained. Therefore, many scholars no longer train SNN parameters directly, but transform the trained parameters in ANN into SNN with the same structure. To achieve near-lossless ANN-SNN conversion, It is necessary to make certain constraints on the original ANN model, such as Different layers of the network are normalized. Sangupta et al weighted and normalized the maximum input received by each layer, which improved the recognition rate of the algorithm in classification tasks. Sinivasan et al proposed a method for transforming ANN into SNN, and first performed constraint training on ANN , including removing the batch normalization layer and bias neurons, then transferring the trained weights from AN to SNN, and finally using the back propagation algorithm based on the approximate derivative of F neurons in the SNN network for training. The advantage of the indirect training algorithm transformed from ANN to SNN is that the weight training method is relatively mature, and the classification effect on traditional data sets is ideal, which can reach the level of deep learning. However, taking constraints on ANN will cause the performance of SNN to decline, and the conversion of SNN training requires a long time step size simulation, and the efficiency is much lower than that of direct training. In addition, the feature extraction step of the indirect training algorithm is completed in the ANN, it is difficult to extract the characteristics of the input information in the time dimension, and it is not suitable for the classification of spatiotemporal data.

Brain-like SMNN shows great potential in processing sparse and discrete data. It can not only process the image information after cotton code, but also mine the characteristics of spatiotemporal data such as speech and EEG from the time dimension. At present, most SN algorithms are still implemented. Based on CPU and GPU processors, such computing platforms with separate data processing modules and storage modules cannot take advantage of SNN's high degree of parallelism and fast computing speed. Therefore, in the past ten years, a series of SN-oriented dedicated hardware computing platforms have emerged, becoming a major branch of brain-inspired computing. The brain-inspired neuromorphic computing platform based on SNN must satisfy: sparse event-driven nature, that is, information transmission is realized in a pulsed manner; it can realize complex dynamic functions, such as realizing the neural nucleus composed of neurons and synapses, and realizing STDP learning rules; A large-scale parallel connection can be realized between the neural cores, and communication can be carried out through a network-on-chip (NoC). The brain-like neuromorphic computing platform mainly completes functions such as information input, weight storage, information weighting, and control pulse distribution by imitating axons, synapses, dendrites, and cell bodies, and realizes the communication between different computing cores by configuring routing functions. data transfer. The existing SNN-like brain neuromorphic computing platforms can be divided into digital-analog hybrid computing platforms and all-digital computing platforms from the circuit technology. Analog circuits can accurately simulate the dynamic characteristics of neurons and realize relatively complex dynamic models. However, analog circuits are easily affected by external factors and have weak programmability. Therefore, many studies tend to use digital-analog hybrid circuits or pure digital circuits. . In the selection of hardware materials, silicon transistors under sub-threshold or super-threshold are usually used, and the implementation technologies include complementary metal-oxide-semiconductor (CMOS) technology and fully depleted silicon technology (fuly depleted-silicon-on-insulator, FDSOI \cite{batude2011de, cheng2016fully, planes201228nm, carter201622nm}) wait. The brain-like neuromorphic computing platform is significantly superior to other hardware systems in terms of volume and energy consumption, and is expected to solve the problems of the failure of Moore's Law and the limitations of the von Neumann system in the future.

\section{Spiking neural network digital-analog hybrid computing platform}

In the digital-analog hybrid computing platform, the analog circuit part can intuitively express the dynamic characteristics of neurons and realize the functions of neurons and synapses; since the routing part needs to complete stable data transmission, it usually adopts a circuit with good stability and high reliability. implemented by digital circuits. Stanford University's Neurogrid system is the most typical digital-analog hybrid brain-like neuromorphic computing platform.
Stanford University's Neurogrid system (11) is a million-level neuron neuromorphic system composed of 16 chips, which consumes only 3.1 W and uses transistors that operate in the subthreshold range. The communication between the 16 chips is through a tree Routing network connections can maximize the number of synaptic connections. Braindrop ("2] is another brain-like neuromorphic chip from Stanford University, which also adopts a digital-analog hybrid design. Compared with the two, the Newrogrid system at the synaptic level Programming requires the use of hardware expertise. Braindop adopts a coupled nonlinear dynamics calculation method and integrates it into the hardware through an automated program, providing a highly abstract programming method to reduce the technical requirements for users. In the future, Stanford will integrate multiple Braindrop cores to build larger Brainstorm chips.
RCL.s("4. Sub-threshold digital-analog hybrid circuits are also used to realize neuron and synaptic dynamics. The network scale is slightly smaller but the bionic effect on the synaptic learning mechanism is better, and the plasticity mechanism based on bistable pulses can be realized. , experience long-term potentiation (ong-term potentiation.LTP) or long-term depression (ong-term depression, LTD). In addition, ROLLs can update synaptic connection strength in real time to realize on-chip online learning, and the energy consumption is only 4 mW.
Different from the above three, BrinScale8 uses a super-threshold digital-horizontal hybrid circuit for neuron dynamics simulation, which can realize short-term inhibition and promotion and STDP two learning rules 4.45). In 2018, the second generation of the BrainScaleS system (Br inScaleS-2 for short) was launched. BrainScaleS-2 uses a complex model that supports nonlinear dendrites and structured neurons, adding a hybrid plasticity scheme4). Compared with the STDP-based fixed learning algorithm in BrinScaleS, the learning algorithm in BrainScaleS-2 can be freely programmed in software and executed on an embedded microprocessor, which can support SNN algorithms and traditional ANN algorithms. 
The DYNAPs neuromorphic processing system and the DYNAP-SEL chip designed by the University of Zurich in Switzerland also use a super-threshold digital-analog hybrid circuit. DYNAPs and DYINAP-SEL adopt a two-level routing scheme to minimize memory usage, combining 2D grid and tree routing, communication between chips through 2D grid routing, and tree routing communication between neural processing cores, and point-to-point Source address routing and combined multicast destination address routing (4349J. This new routing scheme is suitable for the development of emerging storage technologies, such as resistive random-access memory (resistive random-access memory, RAM), phase-change memory (phase -change memory. PCM).
The Neurogrid system uses a dendrite sharing structure and a multicast tree router. Adjacent neurons in the same layer have the same input, and neurons at corresponding positions in different layers have translation-invariant connections, which can maximize throughput, but the Naurogid system does not reflect synapses. The plasticity mechanism cannot be adjusted on-chip. BrinScaleS can implement STDP rules. On this basis, the BrainScaleS-2 system adds a hybrid plasticity solution. Through software-hardware collaboration, on-chip adjustment parameters can be realized. DVNAPS can implement STDP rules and on-chip learning, and choose a combination of layered and mesh routing in the communication scheme, which improves the efficiency of information flow transmission. ROLLS can implement a variety of synaptic plasticity rules and a variety of network structures (feed-forward structure, loop structure), but ROLLS is too small to meet the needs of large-scale networks.

\section{Spiking neural network all-digital computing platform}
The inherent heterogeneity and variability of analog circuits make it difficult to program at the dimension of individual neurons and synapses, but the implementation of all-digital circuits can flexibly adjust the SNN structure and parameters through compatible programming software. On the all-digital computing platform, major companies such as IBM and Intel, as well as top universities such as Manchester University, Tsinghua University, and Zhejiang University have achieved outstanding results.

IBM has been working on neuromorphic processor research since 2008, and has successively produced two achievements, the Goldten Gate chip50) and the TrueNorth processor. In 2018, IBM released the multi-core processor NS16e-4 TrueNorth system, which is composed of TrueNort processors and contains 64 million neurons and 16 billion synapses. The TueNorth series of neuromorphic processors have been applied to various complex tasks, such as: dynamic image recognition in drones or autonomous driving missions, biomedical image segmentation, and EEG signal classification. The SpiNWNaker system at the University of Manchester contains up to 1 036 800 reduced instruction set computer processors (acvanced RISC machine, ARIM)) and 7 terabytes of off-chip dynamic random access memory (dymeamic random access memory DRAM/56.56), The number of neurons that can be simulated is 1\% of the human brain. The team plans to expand the scale of neurons, simulate the entire human brain, and develop the second-generation SpiNaker system (SiNlker2 for short). The spiNaker2 system plans to realize dynamic power management, memory sharing, multiple accumulation accelerators, and neuromorphic accelerators on the basis of the first generation, On-chip network and other functions \cite{haessig2018spiking, palit2018biomedical, kiral2017truenorth}.

Compared to TueNMorth and jSpiNNlaker, ODINV is a small online digital neuromorphic chip that can implement F neurons and 20 different lzhikevich firing patterns56l. MorphIC is the second version of the neuromorphic chip proposed by the team, which is superior to ODIN in terms of scale, and adopts a random version of STDP rules and a hierarchical routing structure, which improves the accuracy of actual tasks Loihi (It is a digital neuromorphic processor released by Intel \cite{davies2018loihi}, which specializes in implementing various synaptic learning rules, not only supports simple pairwise STDP rules, but also supports complex triple STDP rules \cite{blouw2019benchmarking}, reinforcement learning rules with synaptic label assignment, and Using the STDP rule of average rate and pulse timing trajectory (On this basis, the PohoikiBeach system equipped with 64 Lothi chips capable of simulating more than 8.03 million neurons and the PohoikiBeach system equipped with 768 Loth chips capable of simulating more than 100 million Neuron's Pohoiki Springs system has basically taken shape.

The Darwin neural processing unit jointly researched by Zhejiang University and Hangzhou Dianzi University supports a configurable number of neurons, synapses, and synaptic delays, and is a highly configurable neuromorphic chip. Tianji, proposed by Tsinghua University's Brain-Inspired Computing Research Center, is the first heterogeneous fusion neuromorphic computing chip that supports both computer science-based machine learning algorithms and neuroscience-based biologically inspired models. Tianji can freely integrate various neural networks and hybrid encoding schemes, allowing seamless communication between multiple networks (including SNNs and ANs). The team has built on Tianji \cite{deng2020tianjic} for voice command recognition SNN, CNN for image processing and target detection, continuous attractor neural network (continuous attractor neural network.CAN) for human target tracking, long short-term memory network (ong shor-tem menoryLSTM) for natural language recognition And multi-layer perceptron (muirilyer percspron MLP)\cite{pei2019towards} for attitude balance and direction control. Tianji can solve the problem of hardware incompatibility between computational AN and brain-like SNN, and promote the use of SNN in solving practical problems development in. The neural core parameters and connection methods of the TrueNorth neuromorphic processor are highly configurable, and the software-hardware complete correspondence can realize the same program running on the simulator and the chip, but the update of the parameters can only be realized on the software, and cannot be learned on the chip. SpiNaker processor and Loi processor can realize on-chip adjustment of neurons, synaptic parameters and learning rules, especially the processor can configure multiple parameters such as synaptic delay, adaptive threshold, random voice and neuron hierarchical connection. Both the ODIN processor and the Darwin processor only implement a single chip and are small in scale, but the ODIN processor can implement a variety of neuron models, and the density of neurons and synapses is known to be the highest, and the Darwin processor has high configuration. Can meet the needs of practical tasks. Tianji is characterized by the idea of heterogeneous fusion, which can integrate various neural networks and realize communication between different networks.

\section{Conclusion}
This paper summarizes five neuron models commonly used in SNN network construction, namely HH model, IF model, LF model, SHM model and Zhikevich model, and analyzes the circuits, mathematical forms, and advantages and disadvantages of the five models; Network topologies, namely feed-forward SNN, recursive SNN, recurrent SNN, evolutionary SNN and hybrid SNN, summarize the characteristics of the five network topologies. On this basis, the SNN learning algorithm and SNN neuromorphic computing platform are reviewed. First, from the perspectives of unsupervised learning and supervised learning, several thinking directions in the implementation and improvement of SWN algorithm in recent years are summarized; The large-scale SNN neuromorphic computing platform is summarized and analyzed, and the advantages and disadvantages of each computing platform are compared. Through the analysis of current research progress in various aspects of SNN, it can be seen that SNN, as a new generation of neural network, is immature in algorithm and computing platform, and is in the stage of rapid development, facing many challenges, problems and development trends that need to be solved urgently May include the following aspects:

(1) In terms of SMN neuron model and network structure: most of the current SNN networks are based on these five neuron models, especially the LF neuron model. When researchers choose a neuron model, they mainly consider two aspects: one is the calculation amount of the model, and the other is the degree of bionicity of the model. The LF neuron model can best balance these two requirements at present. However, LF neurons only reflect the leakage, accumulation and threshold excitation process of neuron membrane potential, which are much different from the real neuron firing characteristics. Therefore, adding more biological characteristics on the basis of ensuring calculation speed is the future development direction. There are many types of SNN network structures, but in practical applications it is limited to feedforward neural networks. Although the more complex the network may affect the computing efficiency, its role in improving the computing accuracy must still be considered. Therefore, in the network construction, it is necessary to consider adding mechanisms such as loops and feedback.

(2) In terms of SNN learning algorithms: Learning algorithms are the lifeblood of network update iterations. The current application of SNN in the fields of pattern recognition and target detection is far inferior to the unsupervised learning algorithm of traditional ANN based on STDP rules, which can reflect the neurons and synapses in the brain. However, the potential for dealing with large-scale tasks still needs to be explored; several types of supervised learning algorithms mainly start from the perspective of backpropagation and convolution. The accuracy can reach the level of traditional ANN. At present, the SNN algorithm still faces many challenges, which are specifically reflected in: how to apply the learning algorithm based on the STDP rule in the deep network to meet the needs of the recognition task; how to solve the non-differentiable problem of the neuron model based on the back propagation algorithm Solve the problems of over-fitting and robustness; the algorithm for converting ANN to SNN must ensure that the classification accuracy is not lost before and after conversion. In addition, how to truly apply SNN to classification and detection tasks is the most urgent problem to be solved, especially the characteristics of SNN are very suitable for processing spatio-temporal data, such as dynamic visual information, audio and video, EEG, ECG, etc. Potential for dynamic information.

(3) In terms of the SNN neuromorphic computing platform: the SNA neuromorphic computing platform provides new ideas for solving the failure of Moore's Law and the low energy efficiency of the Von Neumann architecture with separation of computing and storage, and is also facing many problems The first is to ensure efficient communication within the neuron core, between neuron cores in a single chip, and between chips. The choice of routing scheme affects the efficiency of information transmission, and an appropriate communication scheme must be selected; the second is to achieve a high degree of configurability of system parameters , most chips have not realized the diversification of configurable parameters, including neuron models, network topology, learning rules and other macro structures, as well as micro adjustments such as synaptic delay and adaptive threshold random noise; the third is to maximize the use of neuromorphic chips in the efficiency and energy consumption advantages to achieve on-chip learning; the fourth is to realize the conversion of on-chip ANN to SNN or parallel computing of ANN and SNN, so that the chip is universal. Overall, SNN is an important inspiration from biological intelligence and will become a class of The basic basis for the realization of brain artificial intelligence. Through the summary of SNN neuromorphic processors, the current processors must cooperate with software to update parameters, and the main learning functions need to be implemented in software. Most processors do not have the ability to handle complex tasks. However, under the rapid development of artificial intelligence, the accuracy of SNN algorithms is gradually improving, and the functions of SNWN-like neuromorphic chips are also gradually increasing. It is expected that SNN will be widely used in pattern recognition, target detection and other fields, The potential in processing spatiotemporal data is gradually tapped.

\clearpage

%--------------------------------------------------

\footnotesize

\bibliographystyle{ieeetr}
\bibliography{references.bib}

\end{document}